\def\year{2022}\relax
\documentclass[letterpaper]{article} % DO NOT CHANGE THIS
\usepackage{aaai22}  % DO NOT CHANGE THIS
\usepackage{times}  % DO NOT CHANGE THIS
\usepackage{helvet}  % DO NOT CHANGE THIS
\usepackage{courier}  % DO NOT CHANGE THIS
\usepackage[hyphens]{url}  % DO NOT CHANGE THIS
\usepackage{graphicx} % DO NOT CHANGE THIS
\usepackage{natbib}  % DO NOT CHANGE THIS AND DO NOT ADD ANY OPTIONS TO IT
\usepackage{caption} % DO NOT CHANGE THIS AND DO NOT ADD ANY OPTIONS TO IT
\DeclareCaptionStyle{ruled}{labelfont=normalfont,labelsep=colon,strut=off} % DO NOT CHANGE THIS
\usepackage{algorithm}
\usepackage{algorithmic}

\usepackage{booktabs} 
\usepackage{bbding}
\usepackage{amsmath}
\usepackage{subfigure} 
\usepackage{amssymb}
\usepackage{amsthm}
\usepackage{color}
\usepackage{makecell}
\usepackage{multirow}

\newtheorem{proposition}{\bf Proposition}

\usepackage{newfloat}
\usepackage{listings}
\floatstyle{ruled}
\newfloat{listing}{tb}{lst}{}
\floatname{listing}{Listing}
\title{STaR: Knowledge Graph Embedding by Scaling, Translation and Rotation}
\author{
    jiayi~Li \textsuperscript{\rm 1}
    Yujiu~Yang \textsuperscript{\rm 1}
}
\affiliations{
    \textsuperscript{\rm 1}Tsinghua Shenzhen International Graduate School, Tsinghua University \\
    lijy20@mails.tsinghua.edu.cn,
}

\begin{document}

\maketitle
\begin{abstract}
The bilinear method is mainstream in Knowledge Graph Embedding (KGE), aiming to learn low-dimensional representations for entities and relations in Knowledge Graph (KG) and complete missing links. Most of the existing works are to find patterns between relationships and effectively model them to accomplish this task. Previous works have mainly discovered 6 important patterns like non-commutativity.  
% 这里可以给出引用，强调其新和重要。另外需要指出对所有模式建模的必要性。
% However, we find that the patterns are not exhausted and discover a pattern non-inversion in contrast to inversion, which essentially associates with 1-to-N, N-to-1, and N-to-N(or complex) relations. 
Although some bilinear methods succeed in modeling these patterns, they neglect to handle 1-to-N, N-to-1, and N-to-N relations (or complex relations) concurrently, which hurts their expressiveness. 
% misunderstand inversion and mistakenly treat some non-invertible relations as invertible ones, which results in the flaw in their models and hurts their performance. To rectify this, we explicitly propose the antithesis of inversion as non-inversion, which essentially associates with 1-to-N, N-to-1, and N-to-N (or complex) relations.
% it指代啥？
% As the new pattern non-inversion emerges, none of the previous models can efficiently model all relation patterns, especially non-inversion and non-commutativity. 
To this end, we integrate scaling, the combination of translation and rotation that can solve complex relations and patterns, respectively, where scaling is a simplification of projection.
% Besides, we know that an arbitrary projection matrix has too many parameters, so we use the simplified scaling matrix instead.
Therefore, we propose a corresponding bilinear model \textbf{S}caling \textbf{T}ranslation \textbf{a}nd \textbf{R}otation (STaR) consisting of the above two parts. Besides, since translation can not be incorporated into the bilinear model directly, we introduce translation matrix as the equivalent. Theoretical analysis proves that STaR is capable of modeling all patterns and handling complex relations simultaneously, and experiments demonstrate its effectiveness on commonly used benchmarks for link prediction. 
% 后面要给出是什么任务上的性能，

%Knowledge Graph Embedding aims to learn low-dimensional representations for entities and relations in Knowledge Graph(KG) and complete the missing links. One important question in this field is what are the patterns among relations and how to model them efficiently. Previous works mainly focus on 4 patterns, which are composition, symmetry, anti-symmetry, and inversion, while two new patterns, commutativity and non-commutativity, are also discovered recently. However, we find that is not complete and discover a pattern non-inversion in contrast to inversion which associates with 1-to-N, N-to-1, and N-to-N(or complex) relations essentially. Based on this discovery, we notice that none of the previous bilinear models is capable to model all relation patterns especially non-inversion and non-commutativity together efficiently. To this end, we notice that in previous works, projection and the combination of translation and rotation are capable to solve the non-inversion and non-commutativity respectively. Besides, we aware that an arbitrary projection matrix has too many parameters, so we use the simplified scaling matrix instead. Therefore, we propose a bilinear model  \textbf{S}caling \textbf{T}anslation \textbf{a}nd \textbf{R}otation(STaR) consisting of the above two parts. STaR is proved to model all 7 patterns and achieves state-of-the-art on commonly used benchmarks. 
%
% However, we discover a neglected pattern: \textit{non-inversion} from another problem that how to deal with 1-to-N, N-to-1, and N-to-N(or complex) relations. This pattern is . 
\end{abstract}

\section{Introduction}
Knowledge Graph (KG), storing data as triples like (head entity, relation, tail entity),  is a growing way to deal with relational data. It has attracted the attention of researchers in recent years due to its applications in boosting other fields such as question answering~\cite{QA}, recommender systems~\cite{recommend}, and natural language processing~\cite{nlp,survey1}.

Since KG is usually incomplete, it needs to be completed by predicting the missing edges. 
A popular and effective way to accomplish this task is Knowledge Graph Embedding (KGE), which aims to find appropriate low-dimensional representations for entities and relations.
% Knowledge Graph Embedding (KGE) is a popular way to accomplish this task, which aims to find appropriate low dimension representations for entities and relations.
%YJ: To solve this issue, knowledge graph embedding (KGE), as a popular method, is utilized to find a suitable low dimension representation for entities and relationships.

% There are two important problems in KGE. One is how to deal with 1-to-N, N-to-1, and N-to-N relations(or complex relations)\cite{transh,transg,transr}. While the other is what are the patterns within the relations of KGs and how to model them efficiently\cite{sun2018rotate,complex,simple}.
% To handle this, it is needed to find patterns within the relations in KGs and model them efficiently~\cite{sun2018rotate,complex,simple}.
 \begin{figure}[ht]
    \centering
    \subfigure[Complex relations and non-commutativity pattern]{
    \includegraphics[width=7.5cm]{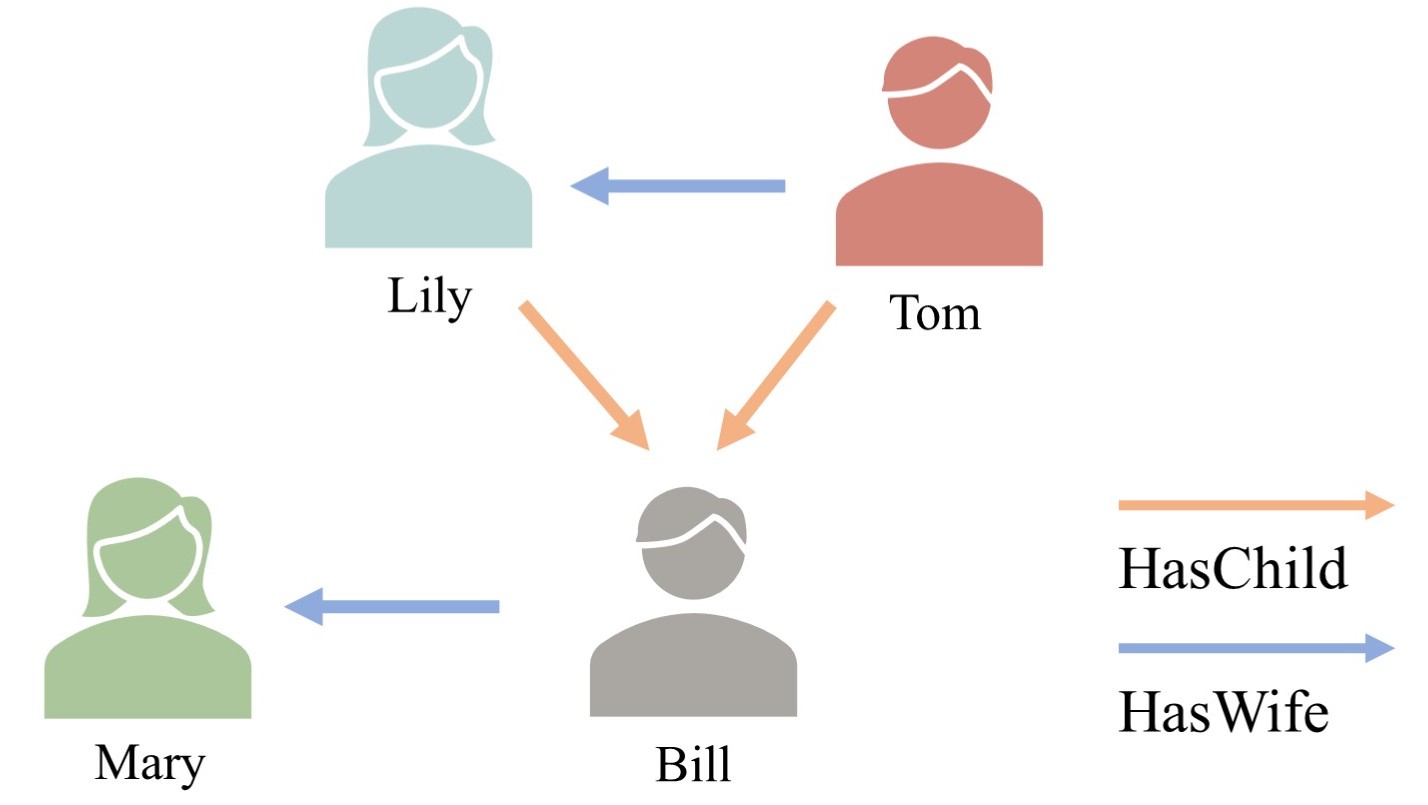}
    \label{fig:example}
    %\caption{fig1}
    }
    \quad
    \subfigure[Translate then rotate]{
    \includegraphics[width=3.5cm]{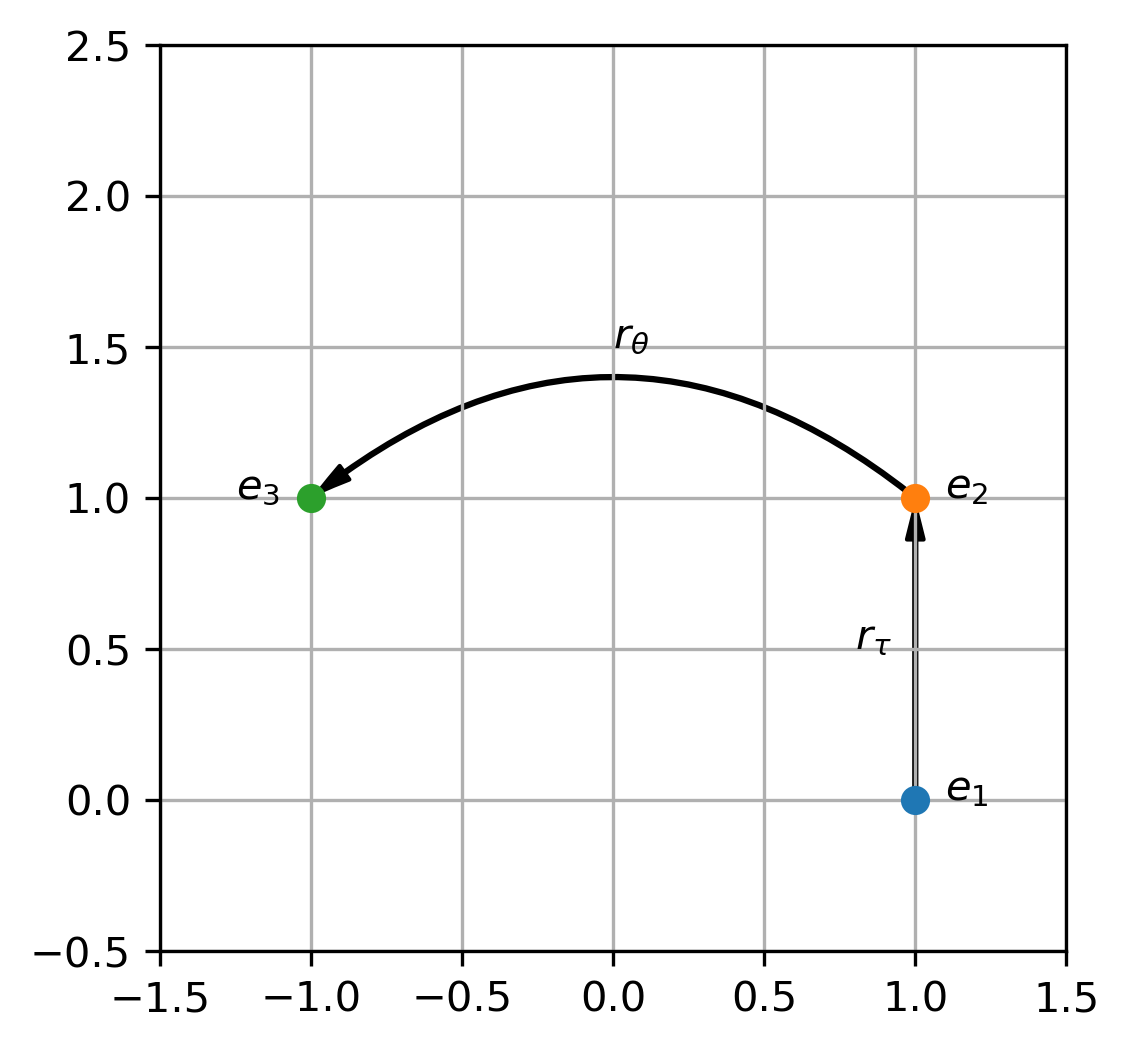}
    \label{fig:TransRotate}
    }
    \subfigure[Rotate then translate]{
    \includegraphics[width=3.5cm]{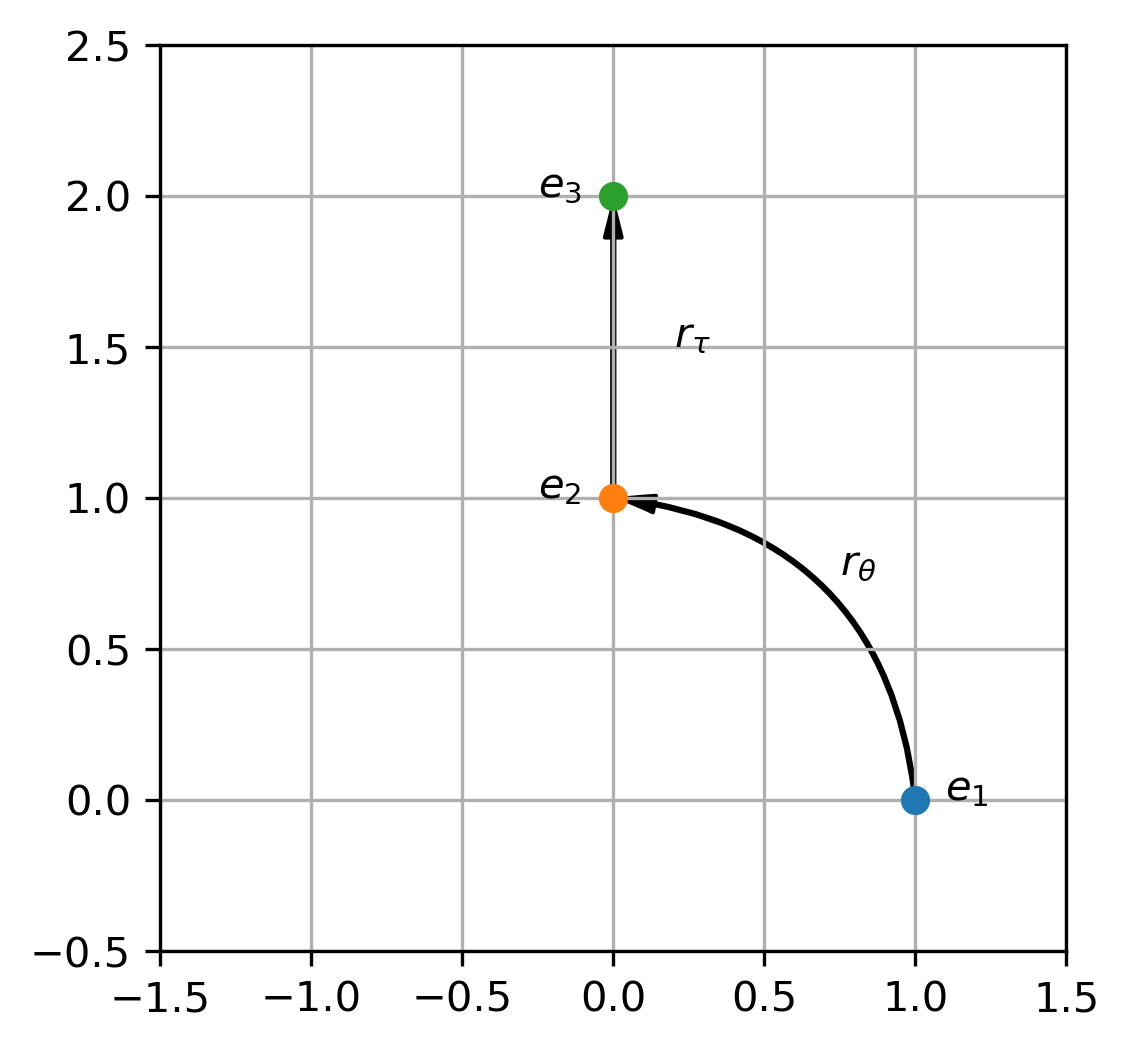}
    \label{fig:RotateTrans}
    }
    \caption{Figure \ref{fig:example} gives examples of N-to-1 relation and non-commutativity pattern. Figure \ref{fig:TransRotate} and Figure \ref{fig:RotateTrans} show how translation and rotation model non-commutativity patterns.}
\end{figure}
% For example, as shown in Fig~\ref{fig:example}, Tom is the father of Mary, and Mary can not be the father of Tom, then the relation \textit{isFatherOf} forms an anti-symmetry pattern. Till now, previous works have discovered 6 important patterns, including non-commutativity and inversion~\cite{NagE}.

% Since diverse and complex relations in KGS, distinguishing and modeling difference patterns is helpful for predicting missing links

% There are two key questions in KGE. One is how to model different relation patterns efficiently. For example, Tom is the father of Mary, then Mary can not be the father of Tom, then the relation \textit{isFatherOf} has an anti-symmetry pattern. Since diverse and complex relations in KGS, distinguishing and modeling difference relation patterns is helpful for predicting missing links. While the other one, as early works mainly focus on, is how to deal with 1-to-N, N-to-1, and N-to-N relations(or complex relations)\cite{transh,transg,transr}.

% One of those patterns is inversion, proposed by RotatE\cite{sun2018rotate}.  For two relations $r_1$ and $r_2$, an inversion pattern is formed by them if for $\forall x,y$, $x\stackrel{r_1}{\longrightarrow}y$ then  $y\stackrel{r_2}{\longrightarrow}x$ and vice versa.

% However, in contrast to the inversion pattern~\cite{sun2018rotate}, we find that previous works neglect the important and prevalent opposite pattern \textit{non-inversion}, 

A mainstream of KGE is the bilinear method, which uses the product of entities and relations as a similarity. While two major problems in KGE are how to model different relation patterns and how to handle 1-to-N, N-to-1, and N-to-N relations (or complex relations)~\cite{sun2018rotate, transh,transr}. For the first problem, previous studies have mainly discovered 6 patterns~\cite{sun2018rotate,dihedral,NagE}. For example, as shown in Figure~\ref{fig:example}, \textit{HasChild} and  \textit{HasWife} form a non-commutativity pattern, since the child of Tom's wife is Bill while the wife of Tom's child is Mary.
For the second problem, we take an N-to-1 relation \textit{HasChild} as an example illustrated in the same figure, in which Bill is the child of both Lily and Tom.

Although some recent works have successfully modeled different relation patterns, they neglect to handle complex relations concurrently. To be more specific, they represent relations as rotations (or reflections) to model different patterns like DihEdral~\cite{dihedral} and QuatE~\cite{quate}, yet ignore that naive rotation is just like translation in TransE\cite{bordes2013translating}, which is difficult to handle complex relations.

To this end, we borrow the ideas from distance-based methods to go beyond rotation and solve these two problems simultaneously. Specifically, we combine projection that handles complex relations~\cite{transh,transr} and the combination of translation and rotation that models relation patterns~\cite{rote}, e.g. we demonstrate how they model non-commutativity in Figure~\ref{fig:TransRotate} and Figure~\ref{fig:RotateTrans}. Thus, we propose a corresponding bilinear model \textbf{S}caling \textbf{T}ranslation \textbf{a}nd \textbf{R}otation (STaR), where scaling is a simplification of projection and translation is introduced as matrix widely used in Robotics~\cite{paul1981robot}. STaR can model different patterns and handle complex relations concurrently, and takes linear rather than quadratic parameters to embed a relation efficiently. Comparing to previous bilinear models, STaR is closest to ComplEx~\cite{complex}, which is equivalent to the combination of rotation and scaling and will be compared in Section \ref{sec:analysis} minutely.
% To solve this, we propose a model \textbf{C}omplEx \textbf{a}nd \textbf{T}ransaltion(\textbf*), which incorporates a translation matrix into ComplEx\cite{complex}, where the ComplEx  matrix is consisted of blocks of 2D-rotation and scaling. STaR makes up the flaw of ComplEx that it is incapable to model non-commutativity proposed by NagE\cite{NagE}, since the combination of rotation and translation in 2D are non-commutative, as shown in Figure \ref{fig:TransRotate} and Figure \ref{fig:RotateTrans}. Thus for n-dimensional embedding, STaR needs $2n$ parameters to embed a relation efficiently rather than $n^2$ in RESCAL. 

% STaR makes up this shortcoming by simply incorporate translation

Experiments on different settings demonstrate the effectiveness of our model against previous ones, while elaborated analysis against ComplEx shows the changes brought about by translation and verifies that our model improves from modeling the non-commutativity pattern.
The main contributions of this paper are as follows:
\begin{enumerate}
    % \item We find the connection between the problems of relationship patterns and complex relations. Then, we subsume the latter problem into the former one by discovering the non-inversion pattern within complex relations.
    \item We propose a bilinear model STaR that can efficiently model relation patterns and handle complex relations concurrently.
    % from the complex relations and thus find the connection between the problems of relation patterns and complex relations.
    % 这句话要改！语义比较含糊。
    \item  To the best of our knowledge, this is the first work introducing translation to the bilinear model, which brings new modules to it and connect with distance-based methods.
    % \item To the best of our knowledge, this is the first paper using translation matrix in bilinear models to simulate the function of translation. Thus a new connection is established between bilinear based models and translational based models.
    % \item We show that Monoid is a better algebra structure than Group for relations in KG. 
    % \item We provide a new perspective that understanding KGs from relation pairs and give corresponding analysis to three KGs commonly used in KGE.
    \item The proposed STaR achieves comparable results on three commonly used benchmarks for link prediction.  
\end{enumerate}

\section{Related Work}
%YJ:这一节有点冗长，需要压缩，并自然过渡到本文解决的问题。
Generally speaking, previous works on KGE can be divided into bilinear, distance-based, and other methods. 

% Here we use $(h, r, t)$ and $s(\cdot)$ to denote the head entity, relation, and tail entity of a specific triple and the score function, respectively.

% Although the above combination of rotation and translation is very powerful, and it is still invertible and thus incapable of modeling the non-inversion pattern.

\subsection*{Bilinear Methods}
Bilinear Methods measure the similarity of head and tail entities by their inner product under a relation specific transformation represented by a matrix $R$. RESCAL~\cite{rescal} is the ancestor of all bilinear models, whose $R$ is arbitrary and has $n^2$ parameters. RESCAL is expressive yet ponderous and tends to overfit. To alleviate this issue, DistMult~\cite{distmult} uses diagonal matrices and reduces $n^2$ to $n$. ComplEx~\cite{complex} transforms DistMult into complex spaces to model anti-symmetry pattern. Analogy~\cite{analogy} considers analogical pattern, which is equivalent to commutativity pattern, and generalizes DistMult, ComplEx, and HolE~\cite{hole}. Although these descendants are powerful in handling complex relations and some patterns, they fail to model non-commutativity patterns.

% ComplEx\cite{complex} finds DistMult is incompetent to model the anti-symmetry pattern and transforms the DistMult into a complex space to resolve this problem.

The non-commutativity pattern was proposed by DihEdral~\cite{dihedral} which uses a dihedral group to model all patterns. Besides, this pattern can also be modeled by hypercomplex values like quaternion or octonion used in QuatE~\cite{quate}. Although they succeed in modeling non-commutativity, they are poor at handling complex relations. Thus, none of the previous bilinear methods has intended to handle relations and model concurrently.
% In addition to ComplEx, some other researchers attempt to model in hypercomplex value. QuatE~\cite{quate} uses quaternion, while GeomE~\cite{geometric} further utilizes geometric algebra to subsume previous works of real value, complex value, and hypercomplex value. Although some succeed in modeling the non-commutativity, they all adopt rotation, which is incapable to handle complex relations. 

% Regardless of the evolution of models, some researchers also investigate how to alleviate the overfitting problem by changing the regularization term. N3\cite{n3} using (...), DURA\cite{dura}

\subsection*{Distance-Based Methods}
In contrast, Distance-Based Methods use distance to measure the similarity.  
TransE~\cite{bordes2013translating} inspired by word2vec~\cite{Mikolov2013DistributedRO} proposes the first distance-based model and model relation as translation. TransH~\cite{transh}, TransR~\cite{transr} find that TransE is incapable to model complex relations like \textit{part\_of} and fix this problem by projecting entities into relation-specific hyperspaces. 

RotatE~\cite{sun2018rotate} utilizes rotation to model inversion and other patterns. Due to its success, subsequent models adopt the idea of rotation. HAKE~\cite{hake} argues that rotation is incompetent to model hierarchical structures and introduces a radial part. MuRE~\cite{murp} incorporates rotation with scaling while RotE~\cite{rote} combines rotation and translation. Besides, they also have hyperbolic versions as MuRP and RotH. PairRE~\cite{pairre} also tries to model both the problems of patterns and complexity together, yet neglects the non-commutativity pattern.

\subsection*{Other Methods}
Apart from the above two, some studies also employ black boxes or external information. ConvE~\cite{conve} and ConKB~\cite{convkb} utilize convolution neural network while R-GCN~\cite{rgcn} and RGHAT~\cite{rghat} apply graph neural networks. Besides, some other works use external information like text~\cite{text,kgbert}, while they are out of our consideration.

Besides specific models, other researchers believe that some previous models are limited by overfitting. Thus, they propose better regularization terms like N3~\cite{n3} and DURA~\cite{dura} to handle this problem.

% \subsection*{Discussion about Group}
% Beyond specific models, some works have revolved around group. TorusE\cite{toruse}, Dihedral\cite{dihedral} uses the properties of certain groups to form their model. And NagE\cite{NagE} tries to give an unified framework under group.

%  For example, It treats RotatE\cite{sun2018rotate} as a concatenation of 1-dim unitary transformation groups.

% However, their discussion ignores that some relations are invertible, which conflicts with the definition of group.
% models ablitiy

% \section{Background Knowledge}
% \label{sec:backgound knowledge}

% \subsection{Abstract Algebra}
% \label{subsec:algebra}
% Here we introduce Group and Monoid. In the Section \ref{section:method} we will discuss why monoid is a better structure for relations.
% \begin{definition}
% \label{def:monoid}
% \cite{jacobson2012basic}. A monoid is a triple $(M, p, 1$ in which $M$ is a non-vacuous set, $p$ is an associative binary composition (or product) in $M$, and 1 is an element of $M$ such that $p(1,a) = a= p(a,1)$ for all $ a \in M$
% \end{definition}

% \begin{definition}
% \label{def:group}
% \cite{jacobson2012basic}. A group $G$ ( or (G,p,1)) is a monoid all of whose elements are invertible.
% \end{definition}

\begin{table*}
\centering
\caption{The score function and ability to model relation patterns of several models.}
\resizebox{\textwidth}{!}{
\begin{tabular}{c|c|cccccc|c}
\hline
\multicolumn{1}{c|}{} &
\multicolumn{1}{c|}{} &
\multicolumn{6}{c|}{\textbf{Relation Patterns}} &
\multicolumn{1}{c}{Performance on} \\
Model & Score Function & Composition & Symmetry & Anti-Symmetry & Commutativity &  Non-Commutativity & Inversion & Complex Relations \\ 
\hline
TransE & $-\|h+r-t\|$&\Checkmark & \XSolidBrush & \Checkmark & \Checkmark & \XSolidBrush & \Checkmark & Low  \\
\hline
TransR & $-\|M_rh + r - M_rt\|$ & \Checkmark & \Checkmark & \Checkmark & \Checkmark & \Checkmark & \Checkmark & High \\
\hline
RotatE & $-\|h\circ r -t \|$ & \Checkmark & \Checkmark & \Checkmark & \Checkmark & \XSolidBrush & \Checkmark & Low\\
\hline
MuRE & $-\|\rho\circ h + r - t \|$ &\Checkmark & \Checkmark & \Checkmark & \Checkmark & \Checkmark & \Checkmark & Low\\
\hline
RotE & $-\|h Rot(\theta_r)+r - t\|$ & \Checkmark & \Checkmark & \Checkmark & \Checkmark & \Checkmark & \Checkmark & Low\\
\hline
DistMult & $h^T\text{diag}(r)t$ & \Checkmark & \Checkmark & \XSolidBrush & \Checkmark & \XSolidBrush & \Checkmark & High \\
\hline
ComplEx & $\text{RE}(h^T\text{diag}(r)\bar{t})$ & \Checkmark & \Checkmark & \Checkmark & \Checkmark & \XSolidBrush & \Checkmark & High \\
\hline
QuatE & $Q_h \otimes W_r^\triangleleft \cdot Q_t$ & \Checkmark & \Checkmark & \Checkmark & \Checkmark & \Checkmark & \Checkmark & Low\\
\hline
STaR & $\hat{h}^TR_{*}\hat{t}$ & \Checkmark & \Checkmark & \Checkmark & \Checkmark & \Checkmark & \Checkmark & High\\

\hline
\end{tabular}}
\label{table:relation}
\end{table*}
\section{Methodology}
\label{section:method}

% In this section, we will first introduce the background knowledge. Then we will expose the misunderstanding on inversion in previous works and propose its antithesis non-inversion to rectify it. Besides, we will prove that non-inversion is substantially associated with complex relations. Therefore, we will propose our model STaR and prove it is capable of modeling 6 relation patterns, including non-inversion and non-commutativity. Finally, we will analyze what translation brings to the bilinear model.

In this section, we will first introduce the background knowledge. Then, we will propose our model STaR by combining the useful modules that solve patterns and complex relations. Finally, we will discuss the translation in the bilinear model.

\subsection{Background Knowledge}
\subsubsection{Knowledge graph}
Given an entity set $\mathcal{E}$ and a relation set $\mathcal{R}$, A knowledge graph $\mathcal{T} = {(h_i, r_j, t_k)} \subset \mathcal{E} \times \mathcal{R} \times \mathcal{E}$ is a set of triples, where $h_i$, $r_j$, $t_k$, denotes the head entity, relation and tail entity respectively. The number of entities and relations are indicated by $|\mathcal{E}|$and $|\mathcal{R}|$.

\subsubsection{Problem definition}
Knowledge graph embedding aims to learn a score function $s(h,r,t)$ and the embeddings of entities and relations, which uses the link prediction task to evaluate the performance. Link prediction first splits triples of the knowledge graph $\mathcal{T}$ into train set $\mathcal{T}_{train}$, test set $\mathcal{T}_{test}$ and valid set $\mathcal{T}_{valid}$. Then, for each specific triple in $\mathcal{T}_{test}$, link prediction aims to give the correct entity $tail \in \mathcal{E}$ a lower rank than other candidates given the query $(head, relation, ?)$ or head entity $head \in \mathcal{E}$ given the query $(?, relation, tail)$ by utilizing the score function.

% Besides, we follow \cite{n3} and introduce a reciprocal relation(not equivalent to the inverse relation) $r^*$ for each relation $r$ and turn each head prediction $(?,r,tail)$ into a tail prediction $(tail, r^*, ?)$.  

\subsubsection{Complex relations}
The complex relations are defined by tails per head and heads per tail of a relation $r$ (tphr and hptr)~\cite{transh}. If tphr $>$ 1.5 and hptr $<$ 1.5 then $r$ is 1-to-N while tphr $>$ 1.5 and hptr $>$ 1.5 means $r$ corresponds to N-to-N.

\subsubsection{Relation patterns} Relation patterns are the inherent semantic characteristics of relations, which are helpful to model relations and inference.

Previous works have mainly proposed 6 patterns~\cite{dihedral,NagE}. They are \textbf{Composition} (e.g., my father's brother is my uncle), \textbf{Symmetry} (e.g., \textit{IsSimilarTo}), \textbf{Anti-Symmetry} (e.g., \textit{IsFatherOf}), \textbf{Commutativity}, \textbf{Non-Commutativity} (e.g., \textit{my wife's son is not my son's wife}), \textbf{Inversion}.
% Here we list the ability to represent these patterns of several models in Tabel \ref{table:relation}.
For the formal definition of all patterns, please refer to Supplementary Material \ref{app:relation pattern}.

\subsubsection{Other notations}
We use $h \in \mathbb{R}^{n\times 1}$ and $t \in \mathbb{R}^{n\times 1}$ to denote the embedding of head entity and tail entity respectively, where $n$ is the embedding dimension. And we use $\circ$ to denote the relation composition. For example, if we take $r_1, r_2, r_3\in\mathcal{R}$, and $r_3$ is the composition of $r_1$ and $r_2$ then $r_3 = r_1 \circ r_2$.

\begin{figure*}[htp]
    \centering
    \includegraphics[width=18cm]{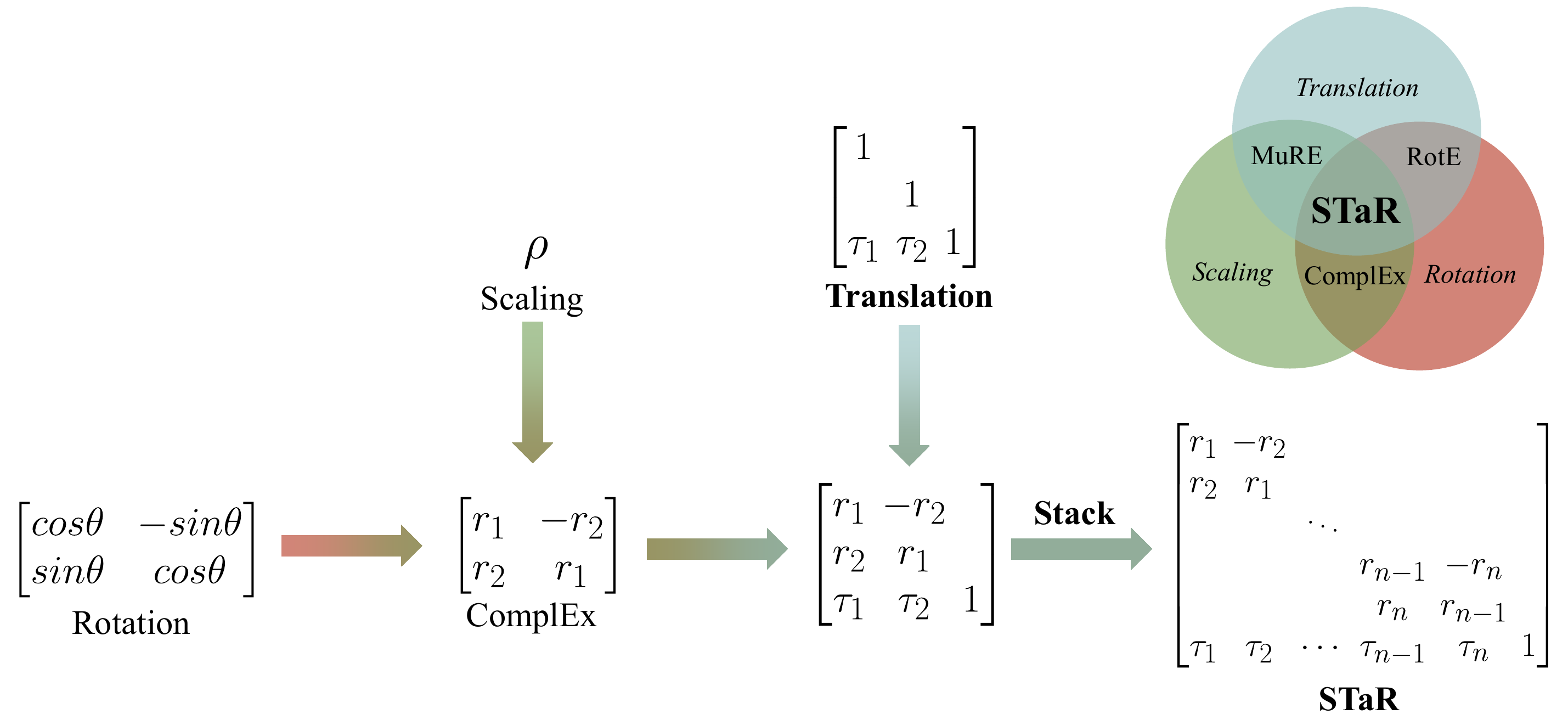}
    \caption{How STaR consists of 3 basic operations and model and related to 3 previous models.} 
    \label{fig:model}
\end{figure*}

% The complex relations are prevalent. Therefore, we can conclude that non-inversion is associated with complex relations substantially and needs to model.

% Apart from the above analysis, we also notice that some previous works misunderstand the idea of inversion and treat some non-invertible relations as invertible. For  instance, in RotatE\cite{sun2018rotate}, it treat hypernym and hyponym as the inversion of each other. But those two relations are complex which means they form the non-inversion rather than inversion pattern based on Proposition \ref{prop:non-invertible}. 

\subsection{The Proposed STaR Model}

In this part, we will analyze modules in previous works that model different patterns and handle complex relations. Then we will propose a bilinear model \textbf{S}caling \textbf{T}ranslation \textbf{a}nd \textbf{R}otation (\textbf{STaR}).

In Table \ref{table:relation}, we list the score function $s(h,r,t)$ of different models and their ability to model patterns, where we observe that the stickiest one is non-commutativity. To model it, QuatE~\cite{quate} utilizes quaternion to model the rotation in 3D space. However, we think it is unnecessary to introduce hypercomplex values and redefine the product operator. In contrast, we are inspired by a distance-based model RotE~\cite{rote} that uses the combination of translation and rotation and is capable of modeling non-commutativity in Euclidean spaces as demonstrated in Figure \ref{fig:TransRotate} and Figure \ref{fig:RotateTrans}.

In the same table, we also list the performance of different models on complex relations. From this table, we notice that scaling, as a special case of projection, is helpful for dealing with complex relations~\cite{distmult,complex}.
% The former can be handled by the combination of rotation and translation in 2Ds as RotE\cite{rote}, which is illustrated in the Figure \ref{fig:TransRotate} and Figure \ref{fig:RotateTrans}. While the latter can be solved by projection as demonstrated in Figure \ref{fig:example}. Besides, projection can be simplified to scaling, which is used by DistMult\cite{distmult} and ComplEx\cite{complex}.

% 加一个to alleviate类的
% Inspired by Robotics~\cite{paul1981robot}, translation is introduced by translation matrix.

Therefore, it seems like we can achieve our goal of modeling patterns and handling complex relations concurrently by assembling the two parts. However, we find that translation is unable to be introduced to the bilinear model directly. To handle this, we introduce a translation matrix widely used in Robotics~\cite{paul1981robot} as the equivalent. To show this substitution, we choose a translation $\tau\in \mathbb{R}^{1\times n}$ to a point $x\in \mathbb{R}^{1\times n}$in $\mathbb{R}^n$ as an example, and we have:
\begin{align}
    \label{equ:trans}
    \begin{bmatrix}x\\1\end{bmatrix}+\begin{bmatrix}\tau \\1\end{bmatrix} = \begin{bmatrix}1 && & \tau_1\\ &\ddots && \vdots \\ && 1 & \tau_n\\ &&&1 \end{bmatrix}\begin{bmatrix}x_1 \\ \vdots\\ x_n \\ 1\end{bmatrix},
\end{align}
where the matrix is the translation matrix.

% On the other hand, we find the ComplEx can be treated as the combination of rotation and scaling in 2Ds, as shown in Fig \ref{fig:model}. 

Finally, we achieve the proposed \textbf{S}caling \textbf{T}anslation \textbf{a}nd \textbf{R}otation (\textbf{STaR}) model by combing such three modules and stacking these elementary blocks as demonstrated in Figure \ref{fig:model}, where ComplEx can be treated as the combination of rotation and scaling in two dimensions manner.

% To solve this question, we choose the scaling matrix as the missing piece of the puzzle. We notice that the scaling matrix is a diagonal matrix and the simplest singular matrix is a diagonal one with some $0$ on its main diagonal. Besides, we find that the combination of rotation and translation in 2D is capable of modeling non-commutativity as illustrated in Figure \ref{fig:TransRotate} and Figure \ref{fig:RotateTrans}. Thus we believe the simplest block to model all 7 patterns is the combination of scaling, rotation, and translation in a 2D real space. As shown in Table \ref{table:relation}, we notice that some previous works are close to this end. Since ComplEx\cite{complex} consists of rotation and scaling in 2D, we only need to introduce translation to complete it.  Therefore, we call such a combination as \textbf{C}omplEx \textbf{a}nd \textbf{T}ranslation(\textbf{STaR}) and  model the relations by stacking such elementary blocks as demonstrated in Figure \ref{fig:model} .

The representation of a relation is thus achieved by assembling a ComplEx matrix with a translation offset:
\begin{equation}
\begin{aligned}
\label{equ:STaR}
    R_{*} = \begin{bmatrix}
    R_c & \\
    \tau^T & 1\\
    \end{bmatrix},
    % R_{STaR} &= R_c \cdot R_{trans}\\
    % &= \begin{aligned}
    %     \begin{bmatrix} 
    %     r^c_1 & - r^c_2\\
    %     r^c_2 & r^c_1 \\
    %     && \ddots \\
    %     &&& r^c_{n-1} & -r^c_{n}\\
    %     &&& r^c_{n} & r^c_{n-1}\\
    %     \tau_1 & \tau_2& \cdots &\tau_{n-1} &\tau_{n} &1
    %     \end{bmatrix},
    % \end{aligned}
\end{aligned}
\end{equation}
where $R_c\in\mathbb{R}^{n\times n}$ and $\tau \in \mathbb{R}^{n\times 1}$ denotes the relation specific ComplEx matrix and translation offset respectively. Besides $R_c$ is achieved by a vector $r^c \in \mathbb{R}^{n \times 1}$ as:
\begin{equation}
\label{equ:complex}
    \begin{aligned}
        R_c = \begin{bmatrix} 
        r^c_1 & - r^c_2\\
        r^c_2 & r^c_1 \\
        && \ddots \\
        &&& r^c_{n-1} & -r^c_{n}\\
        &&& r^c_{n} & r^c_{n-1}\\
        \end{bmatrix}.
    \end{aligned}
\end{equation}

Therefore, the score function of STaR is:
\begin{equation}
\label{equ:score function}
\begin{aligned}
    s(h,r,t) &= \hat{h}^T R_{*} \hat{t}, \\
\end{aligned}
\end{equation}
where $\hat{h} = [h^T,1]^T$ and $\hat{t} = [t^T,1]^T$.

% Finally, we will show that the proposed STaR is able to model all 7 relation patterns as proved in Appendix \ref{proof:relation pattern}.

From the score function, STaR is proved to model all 6 patterns and handle complex relations as detailed in Supplementary Material \ref{proof:relation pattern}.

\begin{proposition}
\label{prop:relation patterns}
STaR can model Symmetry, Anti-Symmetry, Composition, Inversion, Commutativity, and Non-Commutativity and handle complex relations concurrently. 
\end{proposition}

% For the proof, please refer to .

% Besides, STaR only takes $2n$ parameters for a relation, which means it models all 7 relation patterns efficiently.  

\subsection{Discussions}
In this part, we will detail what does translation brings to bilinear model and how it helps to model the non-commutativity minutely.

\subsubsection{What does translation bring to bilinear model?}
We unfold the score function of STaR in Equation (\ref{equ:score function}):
\begin{equation}
\label{equ:vo}
    \begin{aligned}
    s(h,r,t) &= \hat{h}^TR_{*}\hat{t} \\
    % &= \hat{h}^T(R_c\cdot R_{trans})\hat{t}\\
    &=(h^TR_c + \tau^T)t + 1\\
    &=\underbrace{h^TR_ct}_{\text{ComplEx}} + \underbrace{\tau^Tt}_{\text{E}} + 1.
    \end{aligned}
\end{equation}
Except the constant $1$ comes from the extra dimension, the above equation shows that it has two parts: ComplEx and the model E proposed by \cite{fb237}. The later part E is the dot product of the relation-specific translation $\tau$ and the candidate tail entity $t$ regardless of the head entity. Therefore, E works like determining whether the tail entity suits the relation. For example, given a relation \textit{IsLocatedIn}, it is impossible to be a correct triple with a tail entity like \textit{Bill} or \textit{Mary} no matter what the head entity is.  

% As this part functions like choose the appropriate objectives of a certain relation predicate, we call this part Predicate and Object.
% PO works as determining whether the tail entity itself is suitable for this relation regardless of the head entity. 

% Besides,  the symmetrical part $\tau_h^Th$ for the head entity is excrescent in the reciprocal setting that turns head prediction into tail prediction~\cite{n3}. For the tail prediction, $\tau_h^Th$ are fixed, giving the head entity and relation as query, the embeddings of head entity $h$ and translation for head $\tau_h$, thus $\tau_h^Th$ is just a constant. Therefore, this part is unhelpful, which in turn illustrates the simplicity of our model.

\subsubsection{How does translation help model non-commutativity?}
We take two relations $r_1, r_2 \in \mathcal{R}$, whose composited relation $r_3 = r_1 \circ r_2$ is represented as $R_{*}^1\cdot R_{*}^2$. Similarly, we unfold the score function of a triple regarding $r_3$ as:
\begin{equation}
\label{equ:compositon}
\begin{aligned}
    s(h,r_3,t)=& \hat{h}^TR_{*}^1 \cdot R_{*}^2 \hat{t}\\
    =&\left((h^TR_c^1 + (\tau^{1})^T)\cdot R_c^2 + (\tau^2)^T \right) t + 1\\
    =& \underbrace{h^T(R_c^1R_c^2)t}_{\text{ComplEx}} + \underbrace{\left((\tau^1)^TR_c^2 + (\tau^2)^T\right) t}_{\text{E}} + 1. \\
\end{aligned}
\end{equation}
The E in Equation (\ref{equ:vo}) reappears in Equ.(\ref{equ:compositon}). As shown in the Table \ref{table:relation}, it is E, \textit{per se}, helps ComplEx to model the non-commutativity pattern since $(\tau^1)^TR_c^2 + (\tau^2)^T \neq (\tau^2)^TR_c^1 + (\tau^1)^T$.

To better understand the role of E, we take $r_1$ as \textit{IsWifeOf} and $r_2$ as \textit{IsFatherOf} as an example. Then the wife of someone's father must be a woman, while the father of someone's wife must be a man, where the order of relations affects which tail entities are fitted.

\section{Experiments}
In this section, we will introduce the experiment settings and three benchmark datasets and show the comparable results of our model.

\begin{table*}[ht]
\centering
\caption{Link prediction results on different benchmarks (best for $n \in \{200,400,500\}$ . \dag ~means the results are taken from ~\cite{rote}. Since original paper of DURA~\cite{dura} conduct on extremely high dimension, here we reimplement ComlEx-DURA and RESCAL-DURA. Best results are in \textbf{bold} while the seconds are \underline{underlined}. STaR is our full model while TaR excludes scaling.}
\resizebox{\textwidth}{!}{
\begin{tabular}{lcccccccccccc}
\toprule
\multicolumn{1}{c}{} & \multicolumn{4}{c}{\textbf{WN18RR}}& \multicolumn{4}{c}{\textbf{FB15K237}}& \multicolumn{4}{c}{\textbf{YAGO3-10}} \\

Model & MRR & Hits@1 & Hits@3 & Hits@10 & MRR & Hits@1 & Hits@3 & Hits@10 & MRR & Hits@1 & Hits@3 & Hits@10  \\ \midrule

DistMult\dag & 0.43 & 0.39 & 0.44 & 0.49 & 0.241 & 0.155 & 0.263 & 0.419 & 0.34 & 0.24 & 0.38 & 0.54\\ 
ConvE\dag & 0.43 & 0.40 & 0.44 & 0.52 & 0.325 & 0.237 & 0.356 & 0.501 & 0.44 & 0.35 & 0.49 & 0.62 \\
TuckER\dag & 0.470 & 0.443 & 0.482 & 0.526 & 0.358 & 0.266 & 0.394 & 0.544 & - & - & - & -\\
QuatE\dag & 0.488 & 0.438 & 0.508 & 0.582 & 0.348 & 0.248 & 0.382 & 0.550 & - & - & -& -\\
RotatE\dag & 0.476  & 0.428 & 0.492 & 0.571 & 0.338 & 0.241 & 0.375 & 0.533 & 0.495 & 0.402 & 0.550 & 0.670\\
MurP\dag & 0.481 & 0.440 & 0.495 & 0.566 & 0.335 & 0.243 & 0.367 & 0.518 & 0.354 & 0.249 & 0.400 & 0.567\\
RotE\dag & 0.494 & 0.446 & \underline{0.512} & \underline{0.585} & 0.346 & 0.251 & 0.381 & 0.538 & 0.574 & 0.498 & \underline{0.621} & \underline{0.711} \\
RotH\dag & \underline{0.496} & \underline{0.449} & \textbf{0.514} & \textbf{0.586} & 0.344 & 0.246 & 0.380 & 0.535 & 0.570 & 0.495 & 0.612 & 0.706\\
% HAKE$\Diamond$ & 0.497 & 0.452 & 0.516 & 0.582 & 0.346 & 0.250 & 0.381 & 0.542 & 0.545 & 0.462 & 0.596 & 0.694 \\
\midrule
ComplEx-N3\dag & 0.480 & 0.435 & 0.495 & 0.572 & 0.357 & 0.264 & 0.392 & 0.547 & 0.569 & 0.498 & 0.609 & 0.701\\
ComplEx-Fro & 0.457 & 0.427 & 0.469 & 0.515 & 0.323 & 0.235 & 0.354 & 0.497 & 0.568 & 0.493 & 0.613 & 0.699 \\
TaR-Fro (ours) & 0.470 & 0.438 & 0.481 & 0.532 & 0.325 & 0.239 & 0.356 & 0.501 & 0.567 & 0.494 & 0.610 & 0.699\\
STaR-Fro (ours) & 0.463 & 0.431 & 0.476 & 0.526 & 0.324 & 0.236 & 0.356 & 0.501 & 0.574 & 0.502 & 0.617 & 0.701\\
RESCAL-DURA & \underline{0.496} & \textbf{0.452} & \textbf{0.514} & 0.575 & \textbf{0.370} & \textbf{0.278} & \textbf{0.406} & \underline{0.553} & 0.577 & 0.501 & \underline{0.621} & \underline{0.711} \\
ComplEx-DURA & 0.488 & 0.446 & 0.504 & 0.571 & 0.365 & 0.270 & 0.401 & 0.552 & \underline{0.578} & \underline{0.507} & 0.620 & 0.704\\
TaR-DURA (ours) & 0.488 & 0.446 & 0.503 & 0.567 & 0.351 & 0.257 & 0.387 & 0.539 & \underline{0.578} & 0.506 & \underline{0.621} & 0.707\\
STaR-DURA (ours) & \textbf{0.497} & \textbf{0.452} & \underline{0.512} & \underline{0.583} & \underline{0.368} & \underline{0.273} & \underline{0.405} & \textbf{0.557} & \textbf{0.585} & \textbf{0.513} & \textbf{0.628} & \textbf{0.713}\\
\bottomrule
\end{tabular}}
\label{table:high}
\end{table*}

\begin{table}
    \centering
    \begin{tabular}{cccc}
    \toprule
         & WN18RR & FB15K237 & YAGO3-10\\
         \midrule
         $|\mathcal{E}|$ & 40,943 & 14,541 & 123,182\\
         $|\mathcal{R}|$ & 11 & 237 & 37\\
         Train & 86,835 & 272,115 & 1,079,040\\
         Valid & 3,034 & 17,535 & 5,000 \\
         Test & 3,134 & 20,466 & 5,000 \\
         $\Psi$ & 0.003 & 0.801 & 0.838 \\
    \bottomrule
    \end{tabular}
    \caption{Statistics of three benchmark datasets.}
    \label{tab:benchmark}
\end{table}

\subsection{Experiments Settings}
\subsubsection{Datasets}
We evaluate all models on the three most commonly used datasets, which are WN18RR~\cite{conve}, FB15K237~\cite{fb237} and YAGO3-10~\cite{yago3}. 
WN18RR and FB15K237 are the subsets of WordnNet and Freebase, respectively. They are the more challenging version of the previous WN18 and FB15K that suffer from data leakage~\cite{conve,fb237}. We demonstrate the statistics of these benchmarks in Tabel \ref{tab:benchmark}. In particular, we use $\Psi$ to denote the imbalance ratio of the train set, which will be introduced in Section \ref{subsec:compare with complex}

\subsubsection{Baselines}
We compare our method with previous models, which are DistMult~\cite{distmult}, ConvE~\cite{conve}, Tucker~\cite{balavzevic2019tucker}, QuatE~\cite{quate}, MurP~\cite{murp}, RotE and RotH~\cite{rote} and some previous bilinear models with N3~\cite{n3} and DURA~\cite{dura} regularization terms. Besides, we also propose TaR consisting of \textbf{T}ranslation \textbf{a}nd \textbf{R}otation for comparison.  

\subsubsection{Evaluation metrics}
We use the score functions to rank the correct tail (head) among all possible candidate entities. Following previous works, we use mean reciprocal rank (MRR) and Hits@$K$ as evaluation metrics. MRR is the mean of the reciprocal rank of valid entities, avoiding the problem of mean rank (MR) being sensitive to outliers. Hits@$K$ ($K\in\{1,3,10\}$ measures the proportion of proper entities ranked within the top $K$. Besides, we follow the filtered setting~\cite{bordes2013translating} which ignores those also correct candidates in ranking.     

\subsubsection{Optimization}
Following \cite{n3}, we use the cross-entropy loss and the reciprocal setting that adds a reciprocal relation $\Tilde{r}$ for each relation $r\in \mathcal{R}$ and $(t, \Tilde{r}, h)$ for each triple $(h,r,t)\in\mathcal{T}$:
\begin{equation}
    \begin{aligned}
    \mathcal{L} = -\sum_{(h,r,t) \in \mathcal{T}_{train}}(&{\frac{\exp(s(h,r,t))}{\sum_{t'\in \mathcal{E}}{\exp(s(h,r,t'))}}w(t)}\\
    +&{\frac{\exp(s(t,\Tilde{r},h))}{\sum_{h'\in \mathcal{E}}{\exp(s(t,\Tilde{r}, h'))}}w(h))}\\ 
    +& \lambda\text{Reg}(h,r,t),
    \end{aligned}
\end{equation}
where Reg$(h,r,t)$ denotes the regularization and $w(t)(w(h))$ is the weight for the tail (head) entity: 
\begin{align}
    \label{equ:w(t)}
    w(t) = w_0\frac{\#t}{max\{\#t_i:t_i\in\mathcal{T}_{train}\}}+(1-w_0),
\end{align}
where $w_0$ is a constant for each dataset, $\#t$ represents the count of entity $t$ in the training set~\cite{dura}.

Besides, we use both Frobenius (Fro) and DURA\cite{dura} regularization for better comparison. For the details of DURA for STaR please refer to Supplementary Material~\ref{app:detail}.

\subsubsection{Implementation details}
We search the best results in $n\in\{200,400,500\}$. After searching for hyperparameters, we set the dimension to 500, the learning rate to 0.1 for all datasets, and the batch size to 100 for WN18RR and FB15K237 while 1000 for YAGO3-10. Besides, we choose $w_0 = 0.1$ for WN18RR and $0$ for the others. Moreover, for DURA we use $\lambda=0.1,0.05,0.005$ for WN18RR, FB15K237 and YAGO3-10 respectively, while for Frobinues (Fro) we use $\lambda=0.001$ for all cases. Each result is an average of 5 runs.

\subsection{Main Results}
%  the results in high dimension and low dimension settings, respectively. The high dimensional result demonstrates that our model achieves comparable results with baseline models. The low-dimensional result further illustrates the robustness of our model.
% Here we demonstrate the main results. STaR outperforms ComplEx and TaR constantly in different benchmarks, which corroborates that STaR can model patterns and complex relations concurrently.

As shown in Table \ref{table:high}, STaR achieves comparable results against previous bilinear models. STaR improves more on WN18RR and YAGO3-10 than ComplEx under either Fro or DURA regularization. Moreover, STaR achieves similar results compared to RESCAL under DURA. Yet, STaR only needs $2n$ parameters to model a relation while RESCAL requires $n^2$, which shows the efficiency of our model. Besides, STaR still improves about $1\%$ on YAGO3-10 compared to RESCAL. 

Comparing with the distance-based baselines RotE and RotH~\cite{rote}, STaR outperforms them on FB15K237 and YAGO3-10 significantly and gets similar results on WN18RR. Therefore, STaR is more versatile than those distance-based models, which owes scaling.

Besides, we observe that both translation and scaling require appropriate regularization to show their real effects. On the one hand, comparing with STaR-Fro, TaR-Fro achieves similar or even better results, which seems like scaling is useless. On the other hand, comparing with QuatE, TaR-Fro drops $2$ point in WN18RR and FB15K237, which seems like translation and rotation in 2Ds are less powerful than rotation in 3Ds in QuatE. However, that is not the whole story. When we turn to a more powerful regularization term DURA, on the one hand, TaR-DURA is outperformed by STaR-DURA consistently since scaling helps to handle complex relations as shown in Table~\ref{tab:complex relation compare}. On the other hand, TaR-DURA achieves similar results compared to QuatE as they both model all patterns yet are weak on complex relations. We think this phenomenon is because both scaling and translation lack the inborn normalization like rotation and thus require an appropriate regularization term to prevent overfitting.

\begin{table}[ht]
    
    \centering
    \begin{tabular}{rcccc}
    \toprule
        & 1-to-1 & 1-to-N & N-to-1 & N-to-N \\ 
    \midrule
        TaR-DURA & \textbf{0.965} & 0.248 & 0.206 & 0.943 \\ 
        STaR-DURA & 0.922 & \textbf{0.260} & \textbf{0.226} & 0.943 \\ 
        
    \bottomrule
    \end{tabular}
    \caption{The MRR of STaR-DURA and TaR-DURA on complex relations in WN18RR. Better results are in \textbf{bold}.}
    \label{tab:complex relation compare}
\end{table}

\section{Analysis}
\label{sec:analysis}
In this section, we will further compare STaR with ComplEx. Then we will analyze the benchmark KGs in a new perspective to explain the unexpected phenomenon in the comparison. Finally, we will verify that the improvement comes from modeling non-commutativity. 
\begin{figure}[ht]
    \centering
    \includegraphics[width=0.5\textwidth]{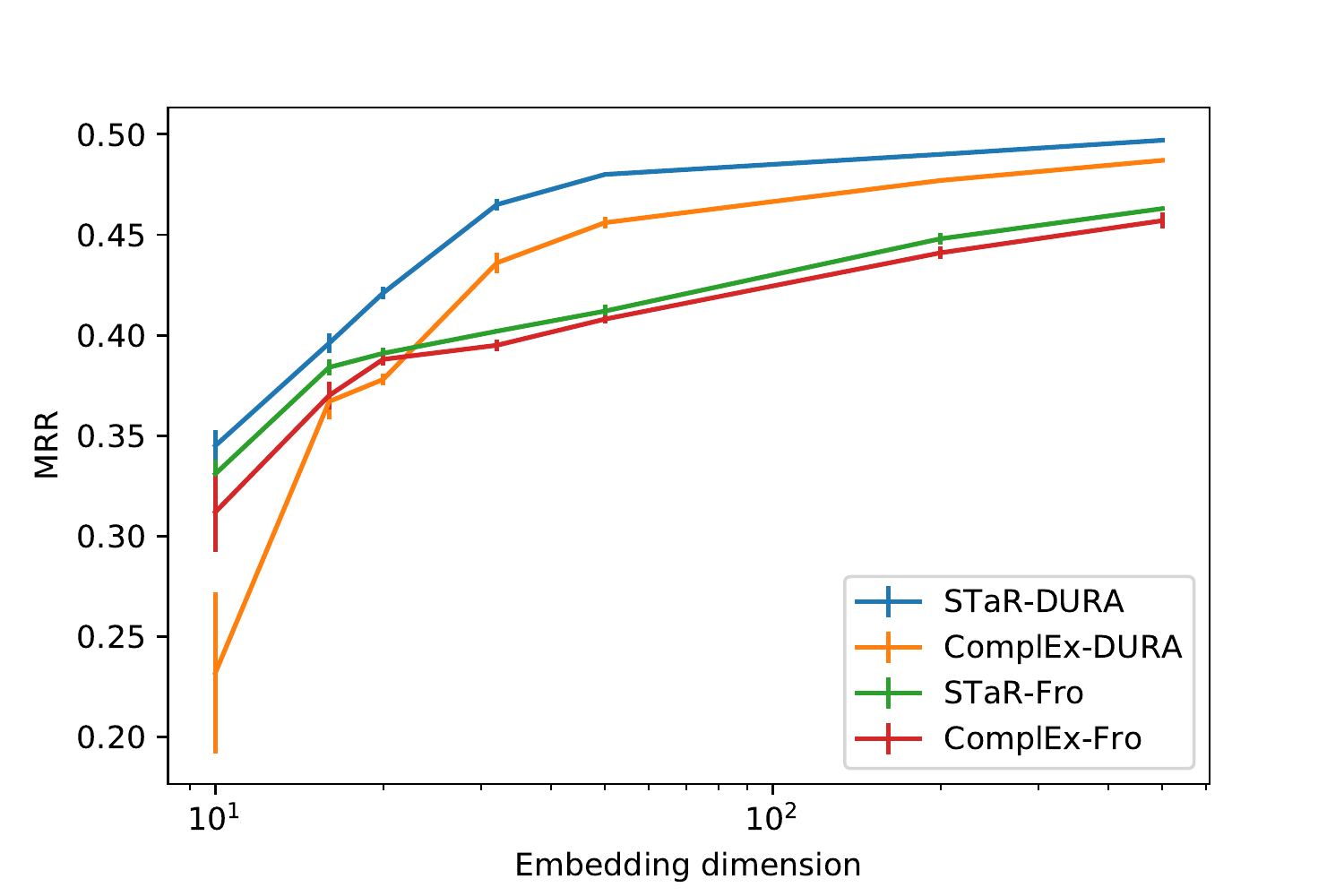}
    \caption{Comparison of STaR and ComplEx on WN18RR under different dimensions ($n \in\{10,16,20,32,50,200,500\}$) and regularization terms (Fro and DURA). Averages and standard deviations are computed over 5 runs for each case.}
    \label{fig:comparasion on wn18rr}
\end{figure}

\subsection{Further Comparison with ComplEx}
\label{subsec:compare with complex}

% In order to measure whether this imbalance of orders exists in those three datasets.
\begin{figure*}[htbp]
\centering
\subfigure[WN18RR]{
\includegraphics[width=5cm]{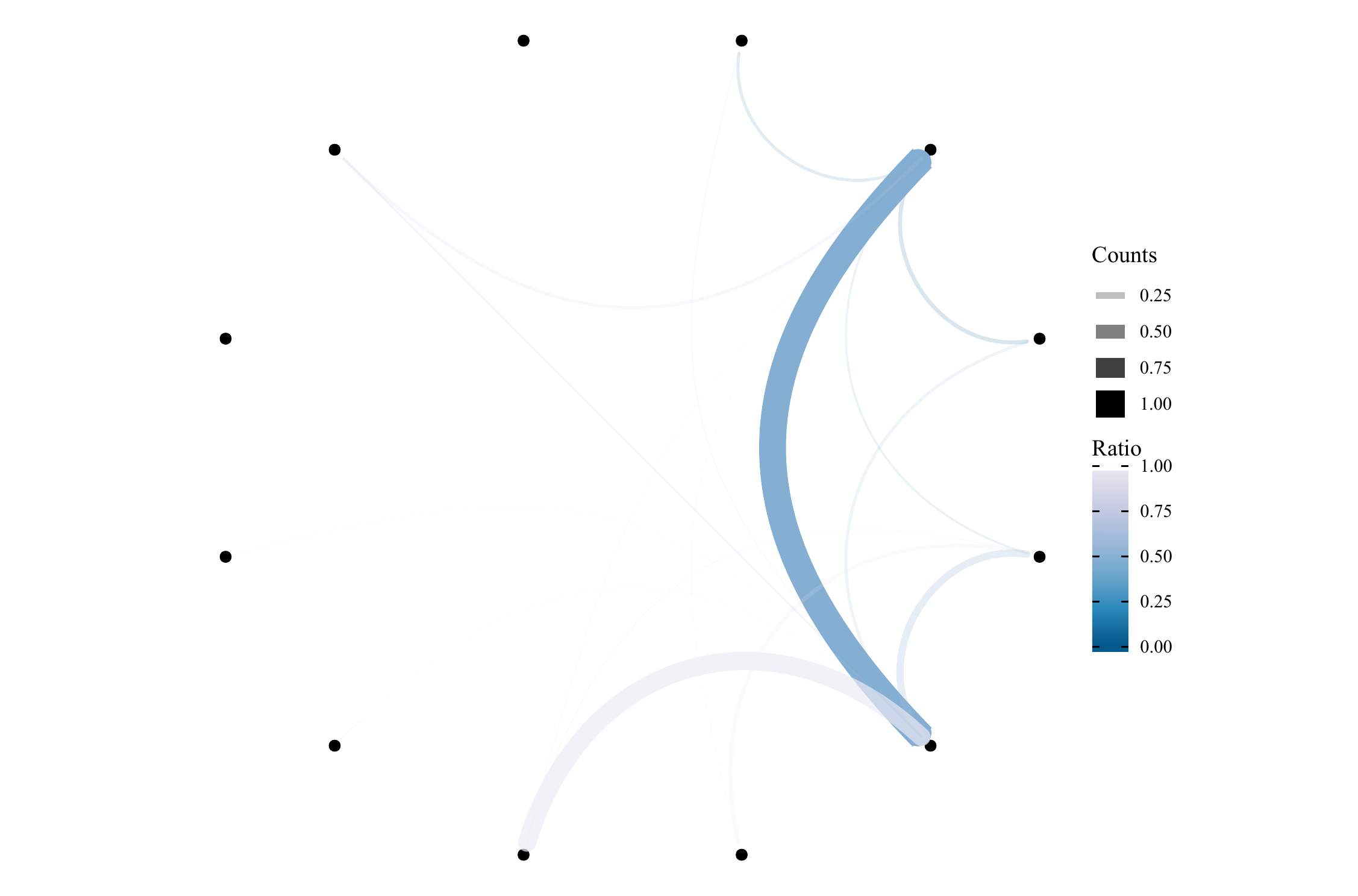}
\label{fig:visual:WN18RR}
%\caption{fig1}
}
\quad
\subfigure[FB15K237]{
\includegraphics[width=5cm]{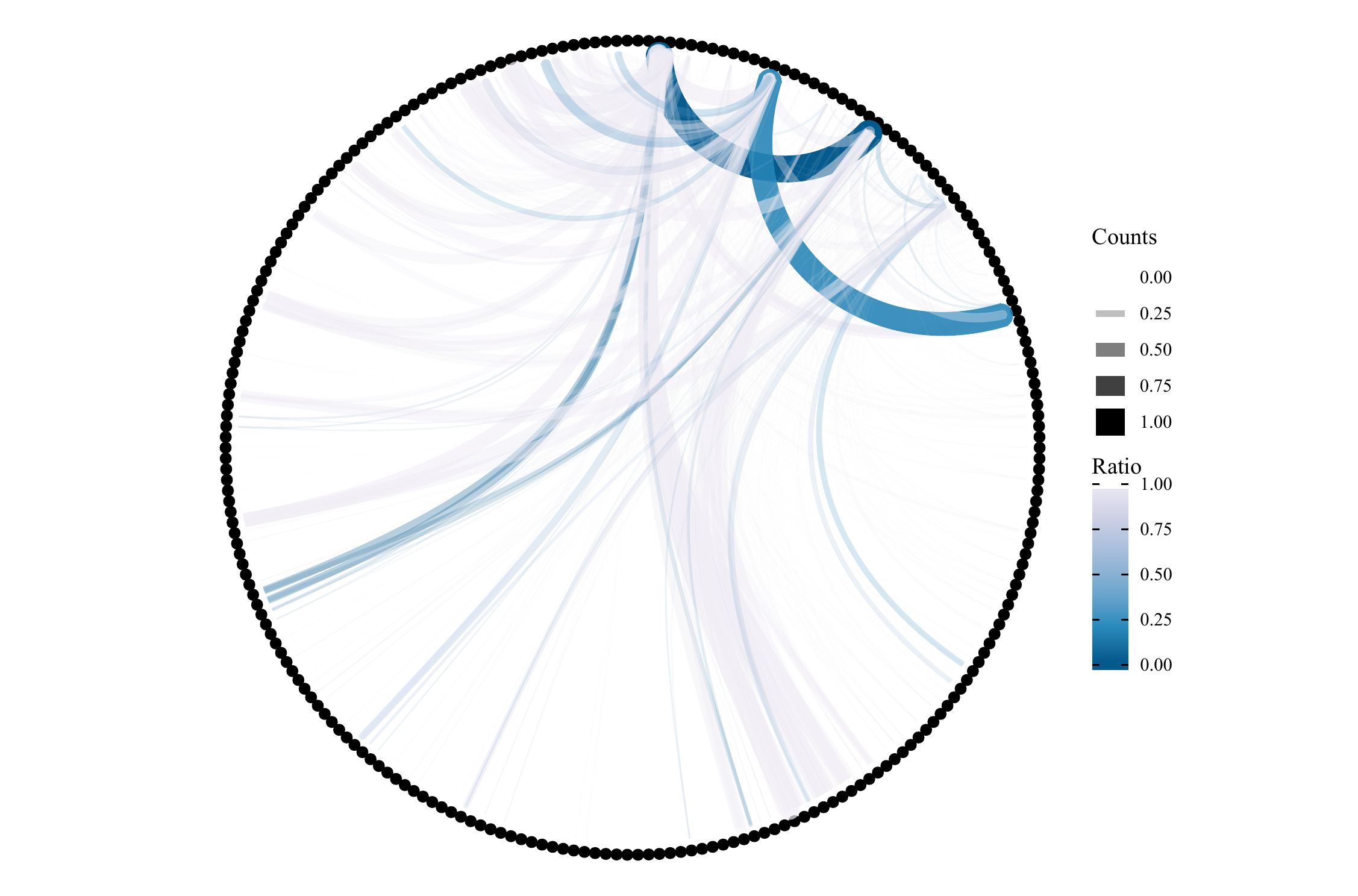}
}
\quad
\subfigure[YAGO3-10]{
\includegraphics[width=5cm]{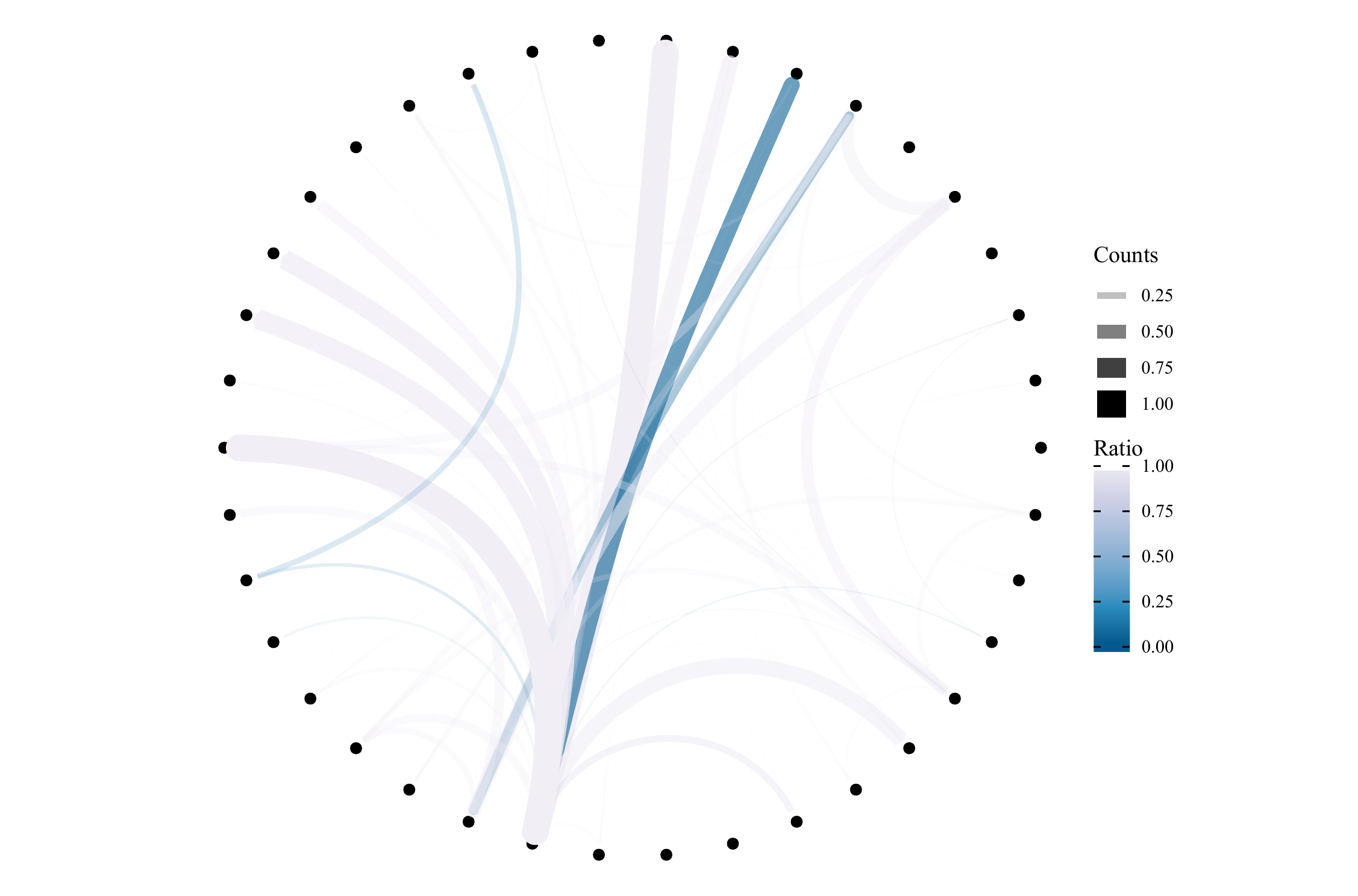}
}
\caption{The count and imbalance ratio of all possible pairs. An arc represents a pair. On the one hand, pair imbalance ratio $\psi$ is denoted by color, as blue means balance while gray means imbalance in contrast. On the other hand, the count is denoted by transparency and thickness, as thick and opaque means more while thin and transparent means less. It should be noticed that the thickness of the arcs is relative, so the arcs with the same thickness in different datasets may have different counts.}
\label{fig:visulization}
\end{figure*}

To show STaR outperforms ComplEx consistently, we conduct further experiments in different dimensions and regularization terms. As shown in Figure \ref{fig:comparasion on wn18rr}, STaR exceeds ComplEx on WN18RR persistently. Besides, both STaR and ComplEx improve by substituting DURA for Frobenius as the dimension increases. Additionally, STaR and ComplEx seem to intersect in an extremely high dimension, which leads us to further experiment in the following content.

\begin{table}
\centering
\caption{Link prediction results between STaR and ComplEx for extremely high-dimensional embedding (best for $n \in \{1000,2000,4000\}$). Better results are in \textbf{bold}.} 
\scalebox{0.7}{
\begin{tabular}{lcccccc}
\toprule
\multicolumn{1}{c}{} & \multicolumn{2}{c}{\textbf{WN18RR}}& \multicolumn{2}{c}{\textbf{FB15K237}}& \multicolumn{2}{c}{\textbf{YAGO3-10}} \\
Model & MRR & Hits@10 & MRR & Hits@10 & MRR & Hits@10  \\ \midrule
ComlEx-DURA & 0.490 & 0.573 & \textbf{0.371} & \textbf{0.561} & 0.583 & 0.710 \\
STaR-DURA & \textbf{0.499}& \textbf{0.585} & 0.370 & 0.558 & \textbf{0.584} & \textbf{0.713}\\
\bottomrule
\end{tabular}}
\label{table:extremely high}
\end{table}

As shown in the Tabel \ref{table:extremely high}, STaR outperforms ComplEx on WN18RR prominently. However, these two are tied on FB15K237 and YAGO3-10 unexpectedly. We think such a phenomenon is due to the lack of non-commutativity patterns in them substantially. To verify our hypothesis, we further investigate those KGs from a new perspective.

\subsection{Imbalance Ratio among KGs}
In this part, we will verify the above hypothesis by introducing two matrices $\psi$ and $\Psi$ about the imbalance ratio.

We find that modeling commutativity and non-commutativity is useful only if both possible orders of a pair of relations appear in a KG. For instance, consider two relations $r_1, r_2 \in \mathcal{R}$, which have two possible orders of composition: $r_1\circ r_2$ and $r_2 \circ r_1$. Therefore, if only one of them, e.g., $r_1 \circ r_2$, exists in the KG, it is unnecessary to distinguish whether they are commutative or not, which we regard as an imbalance.

To this end, we propose two matrices $\psi$ and $\Psi$ to evaluate the imbalance ratio of pair and KG, respectively. For the details of these two matrices, please refer to Supplementary Material \ref{app:matrix}.

% Besides, we list $\Psi$ of each benchmark in Table \ref{tab:benchmark}.

Based on $\Psi$ of each benchmark as shown in Table \ref{tab:benchmark}, we observe that the imbalance is remarkable in FK15K237 and YAGO3-10. Moreover, we are aware that although some pairs have both orders, the counts between orders may have an enormous discrepancy. To show this more specifically, we visualize the pairs of three benchmark KGs. As shown in Figure~\ref{fig:visulization}, on the one hand, the majority of pairs are imbalanced in FB15K237 and YAGO3-10. On the other hand, although many imbalanced pairs exist in WN18RR, the balanced ones account for the majority as denoted by $\Psi$.

% the balanced ones occupy more triples and thus make $\Psi$ small and  valuable to distinguish these two patterns. 

%  On the other hand, although many of the pairs in WN18RR are also imbalanced, the outstanding balanced pair occupied a large proportion of relation pairs, $\%45.5$ to be more specific.

We believe the above analysis validates the hypothesis and explains the phenomenon. Furthermore, we think the discrepancy between KGs is rooted in the entities. Specifically, we notice that all entities are homogeneous in WordNet, which consists of words, while heterogeneous in Freebase and YAGO, built by various things like person, film, etc. Therefore, in KGs like WordNet, all relations connect things of the same kind. In contrast, in ones like Freebase and YAGO, most relations connect things of different kinds.
% which means the combinations of relations need to fit their kinds, and thus most of them have one order solely. 

Therefore, for the relations in the imbalance KGs like FB15K237 and YAGO3-10, some pairs of them only have one meaningful order in the sense of semantics substantially. For instance, consider two relations: \textit{isDirectedBy} and \textit{likeEating}, whose combination makes sense in the order of $film\stackrel{\textit{IsDirectedBy}}{\longrightarrow}human\stackrel{\textit{likeEating}}{\longrightarrow}food$. However, when exchanging the order, we find that the tail entity of \textit{likeEating} should be a kind of food, and the head entity of \textit{isDirectedBy} should be a movie, which shows the inherent incompatibility in this order. More generally speaking, taking $\forall r_1, r_2 \in\mathcal{R}$ that has the order of combination $r_1\circ r_2$. Its other order $r_2\circ r_1$ is meaningless and nonexistent if the domain of head entity of $r_1$ and tail entity of $r_2$ are not intersected. In conclusion, we think that such a semantic character of these inter-kind relations explains the cause of the scarcity of non-commutativity in FB15K237 and YAGO3-10.

% Thus, this relation pair has only one possible order, and we do not need to distinguish commutativity and non-commutativity in this situation at all.

% If the domain of head entity of $r_1$ and the domain of tail entity of $r_2$ is not intersected, then  $r_2\circ r_1$ is meaningless.

\begin{table}
\centering
%\caption{Comparison of the MRR of STaR and ComplEx on WN18RR. $\bigtriangleup$ denotes improvement while $\bigtriangledown$ decrease on extremely high dimensional setting.}
\caption{Comparison of the MRR of STaR and ComplEx on WN18RR. $\bigtriangleup$ denotes improvement and $\bigtriangledown$ decreases on extremely high-dimensional settings.}
\scalebox{0.7}{
\begin{tabular}{lrrrrl}
\toprule
\multicolumn{1}{l}{Relation Name} & \multicolumn{1}{l}{Propotion}& \multicolumn{1}{l}{\textbf{STaR}} & \multicolumn{1}{l}{\textbf{ComplEx}} & \multicolumn{1}{l}{Improvement} &  \\
\midrule
hypernym &40.09\%& 0.193 & 0.175 & $10.29\%\ \bigtriangleup$  &  \\
derivationally related form & 34.23\% & 0.956 & 0.959 & $-0.31\%\ \bigtriangledown $  &  \\
member meronym&8.52\% & 0.241 & 0.225 & $7.11\%\ \bigtriangleup$  &  \\
has part &5.55\%& 0.247 & 0.230 & $7.39\%\ \bigtriangleup$  &  \\
synset domain topic of&3.56\%& 0.409 & 0.387 & $5.68\%\ \bigtriangleup$  &  \\
instance hypernym&3.37\% & 0.420 & 0.409 & $2.69\%\ \bigtriangleup$  &  \\
also see & 1.49\% & 0.634 & 0.631 & $0.47\%\ \bigtriangleup $  &  \\
verb group &1.30\% & 0.917 & 0.975 & $-5.95\%\ \bigtriangledown$ &  \\
member of domain region&1.06\% & 0.408 & 0.279 & $46.24\%\ \bigtriangleup$  &  \\
member of domain usage &0.73\%& 0.359 & 0.316 & $13.61\%\ \bigtriangleup$  &  \\
similar to &0.09\%& 1.000 & 1.000 & $0.00\%\quad\,$            &  \\
\bottomrule
\end{tabular}}
\label{table:complex vs STaR}
\end{table}

\subsection{Improvements on WN18RR Come from Modeling Non-Commutativity Pattern}
In FB15K237 and YAGO3-10, we have shown that imbalances are prevalent and thus explain why STaR and ComplEx are tied. Here we further experiment to corroborate that the improvement on WN18RR gains from modeling the non-commutativity pattern. 

%  From the Table \ref{table:complex vs STaR}, STaR surpasses ComplEx in some relations. Besides, the outstanding thick blue arc in Figure\ref{fig:visual:WN18RR} denotes both $e_1 \stackrel{hyp.}{\longrightarrow} e_2\stackrel{\textit{d.r.f.}}{\longrightarrow}e_3$ and $e_1 \stackrel{\textit{d.f.r.}}{\longrightarrow} e_2\stackrel{\textit{hyp.}}{\longrightarrow}e_3$ are abundant in WN18RR\footnote{\textit{hyp.} and \textit{d.f.r} stands for \textit{hypernym} and \textit{derivationally related form} respectively.}. Besides, it should also be noticed that these two relations are non-commutative.

% Correspondingly, STaR slightly decreases in \textit{derivationally related form} which is already very high and gains about $10\%$ in \textit{hypernym}. Therefore, it validates that our model is able to model non-commutative and thus increases the performance.

As shown in Table \ref{table:complex vs STaR}, STaR surpasses ComplEx in most relations. Although STaR slightly decreases in \textit{derivationally related form} which is already high enough, it gains about $10\%$ in \textit{hypernym} with the largest proportion. Correspondingly, we notice that in Figure \ref{fig:visual:WN18RR} the outstanding thick blue arc denotes both $e_1 \stackrel{hyp.}{\longrightarrow} e_2\stackrel{\textit{d.r.f.}}{\longrightarrow}e_3$ and $e_1 \stackrel{\textit{d.f.r.}}{\longrightarrow} e_2\stackrel{\textit{hyp.}}{\longrightarrow}e_3$ are abundant in WN18RR\footnote{\textit{hyp.} and \textit{d.f.r} stands for \textit{hypernym} and \textit{derivationally related form} respectively.}. Besides, we find that these two relations are non-commutative. Therefore, we think such a correspondence validates that the improvement on WN18RR comes from modeling non-commutativity.

\section{Conclusion}
In this paper, we notice that none of the previous bilinear models can model all patterns and handle complex relations simultaneously. To fill the gap, we propose a bilinear model \textbf{S}caling \textbf{T}ranslation \textbf{a}nd \textbf{R}otation (STaR) consisting of these three basic modules. STaR solves both problems concurrently and achieves comparable results compared to previous baselines. Moreover, we also conduct a deep investigation to verify that our model is improved by handling relations or modeling patterns that previous bilinear models failed.

% and we discover a neglected pattern non-inversion which most of these works fail to model. Besides, we also prove that non-inversion is associated with non-inversion. When considering the new pattern, we find previous bilinear models fail to model all relation patterns efficiently. To this end, we introduce translation by matrix and propose STaR consisting of scaling, translation, and rotation. The experiments show the effectiveness of our model. Further investigation on STaR confirms our improvements compare to previous models.

% Beyond specific models, we also suggest using monoid rather than group as the unified framework for relations and analyzing KGs in the perspective of imbalance, which we will further study in future works.
% We will try to discover more relation patterns and find the deeper connection between the bilinear and translational model in future works. We hope the ideas of translation matrix, monoid, and imbalance relation ratio will inspire the following researches.

% The experiments show the effectiveness of our model. Besides, we suggest that monoid rather than group is a better abstract algebra structure for relations. Finally, we analyze three common benchmarks in the view of relation pairs and uncover that some pairs only exist in one meaningful order substantially. We hope that the new perspective of non-inversion, monoid, and the order can inspire follow-up researches.

\bibliography{KGE.bib}
\clearpage
\appendix
\section*{Supplementary Material}
\section{Formal Definitions of 7 Relation Patterns}
\label{app:relation pattern}
Consider triples of a completed KG $\mathcal{T}^*$, which contains all true facts for entities $\mathcal{E}$ and relations $\mathcal{R}$. Therefore, the former definition of those  patterns are as follows:
\begin{enumerate}
\item \textbf{Symmetry}: For a relation $r\in \mathcal{R}$ and $\forall e_1, e_2\in\mathcal{E}$, if $(e_1, r, e_2)\in\mathcal{T}^*$ then $(e_2, r, e_1)\in\mathcal{T}^*$.
\item \textbf{Anti-Symmetry}: For a relation $r\in \mathcal{R}$ and $\forall e_1, e_2\in\mathcal{E}$, if $(e_1, r, e_2)\in\mathcal{T}^*$ then $(e_2, r, e_1)\notin\mathcal{T}^*$.
\item \textbf{Composition}: For relations $r_1, r_2, r_3\in \mathcal{R}$ and $\forall e_1, e_2, e_3\in\mathcal{E}$, if $(e_1, r_1, e_2)\in\mathcal{T}^* \wedge (e_2, r_2, e_3)\in\mathcal{T}^*$ then $(e_1, r_2, e_3)\notin\mathcal{T}^*$. Therefor, $r_3$ is the composition of $r_1$ and $r_2$.
\item \textbf{Commutativity}: For relations $r_1, r_2 \in \mathcal{R}$ and $\forall e_1, e_2, e_3 \in \mathcal{E}$, if $(e_1, r_1, e_2)\in\mathcal{T}^*\wedge(e_2, r_2, e_3)\in\mathcal{T}^*$ then $(e_1, r_2, e_2)\in\mathcal{T}^*\wedge(e_2, r_1, e_3)\in\mathcal{T}^*$.
\item \textbf{Non-Commutativity}: For relations $r_1, r_2 \in \mathcal{R}$ and $\forall e_1, e_2, e_3\in\mathcal{E}$, if $(e_1, r_1, e_2)\in\mathcal{T}^*\wedge(e_2, r_2, e_3)\in\mathcal{T}^*$ then $(e_1, r_2, e_2)\notin\mathcal{T}^*\vee(e_2, r_1, e_3)\notin\mathcal{T}^*$. 
\item \textbf{Inversion}: For relations $r_1, r_2 \in \mathcal{R}$ and $\forall e_1, e_2\in\mathcal{E}$ if $(e_1, r_1, e_2)\in\mathcal{T}^*$ and $(e_2, r_2, e_1)\in\mathcal{T}^*$  \textit{iff} $e_2 = e_1$.
% \item \textbf{Non-Inversion}:
% For relations $r_1, r_2 \in \mathcal{R}$ and $\forall e_1, e_2\in\mathcal{E}$ if $r_1$ and $r_2$ is not invertible then $r_1$ and $r_2$ is non-invertible, and thus form a non-inversion pattern.
\end{enumerate}

\section{Proof of Proposition \ref{prop:relation patterns}}
\label{proof:relation pattern}
\begin{proof}
Since each relationship is represented by a matrix $R_*$ and the matrix multiplication stands composition operator $\circ$, here we will show how to model all 6 properties by taking some cases of $R_*$ and how to handle complex relations by considering a fixed margin $\gamma$ .
\begin{enumerate}
    \item \textbf{Symmetry}: Here we take $r^c_i = 0, \quad i=1,3,\cdots,n-1$ and $\tau = \mathbf{0}$. Then STaR degenerates to DistMult. Thus $\hat{h}^TR_*\hat{t} = \hat{t}^TR_*\hat{h}$ and STaR models the symmetry pattern.
    \item \textbf{Anti-Symmetry}: Here we take $r^c = \mathbf{0}$ and $\tau \in \mathbb{R}^{n\times 1}$, and STaR degenerates to TransE\cite{bordes2013translating}. Then it models the anti-symmetry pattern, since if $\|h+r-t\| = 0$ then $\|t+r-h\|\neq0$ for $h,r,t\neq \mathbf{0}$. 
    \item \textbf{Composition}: It is equivalent that taking $R_{*}^1, R_{*}^2$ then $R_{*}^1 \cdot R_{*}^2$ is still in the form of $R_{*}$:
    \begin{equation*}
        \begin{aligned}
        R_{*}^1 \cdot R_{*}^2 &= \begin{bmatrix}R_c^1 \\ (\tau^1)^T & 1 \end{bmatrix} \cdot \begin{bmatrix}R_c^2 \\ (\tau^2)^T & 1 \end{bmatrix} \\
        &= \begin{bmatrix}R_c^1 \cdot R_c^2 \\ (\tau^1)^T R_c^2  + (\tau^2)^T& 1 \end{bmatrix},
        \end{aligned}
    \end{equation*}
    thus STaR can model the composition pattern. 
    \item \textbf{Commutativity}: If we take $\tau = \mathbf{0}$, then STaR degenerates to ComplEx matrix, which is a block diagonal matrix and can be exchanged $R_{*}^1 \cdot R_{*}^2  = R_{*}^2 \cdot R_{*}^1$. Thus STaR can model the commutativity pattern.
    \item \textbf{Non-Commutativity}: As demonstrated in Figure \ref{fig:TransRotate} and Figure \ref{fig:RotateTrans}, the translation and rotation are non-commutative. Then, we take $\tau^1 = \mathbf{0}$ and $r_i^2 + r_{i+1}^2 = 1,\quad i=1,3,\cdots,n-1$ to degenerate $R_{*}^1$ into a pure rotation matrix, and $r_c^2 = \mathbf{0}$ to degenerate $R_{*}^2$ into a pure translation matrix. Then, $R_{*}^1 \cdot R_{*}^2 \neq R_{*}^2 \cdot R_{*}^1$  
    \item \textbf{Inversion}: Here we take $\tau = \mathbf{0}$, then for a $R_*^1$, there exists $R_*^2$ that has $(R_*^1)^T = R_*^2$. Therefore, we have $\hat{h}^TR_*^1\hat{t} = \hat{t}^TR_*^2\hat{h}$.
    \item \textbf{Complex relations}
    Here we follow~\cite{pairre} and treat the ability of model handling complex relations is adaptive adjusting the margin given a fixed one. Specifically, we set this fixed margin as $\gamma$, and a candidate is true means the score of the corresponding triple $s(h,r,t)$ is greater than $\gamma$:
    \begin{align}
        \gamma < h^TRt  .
    \end{align}
    If a constant $\alpha$ is multiplied on both side and only changes $R$, then we say it adaptively adjusts the margin. Therefore, for a $(h,r,t)$, STaR has: 
    \begin{equation}
        \begin{aligned}
        &\gamma < \hat{h}^TR_*\hat{t}  \\
        \alpha&\gamma < \alpha\hat{h}^TR_*\hat{t}  \\
        \alpha&\gamma < \hat{h}^T\alpha\begin{bmatrix}
        R_c & \\ & 1
        \end{bmatrix}\begin{bmatrix}
        I \\
        \tau^T & 1
        \end{bmatrix}\hat{t} \\
        \alpha&\gamma < \hat{h}^T\left(\begin{bmatrix}
        \alpha R_c & \\ & 1
        \end{bmatrix} + \begin{bmatrix}
        \mathbf{0}^{n\times n} & \\ & (\alpha - 1)
        \end{bmatrix}\right)\begin{bmatrix}
        I \\
        \tau^T & 1
        \end{bmatrix}\hat{t}\\
        \alpha&\gamma < \hat{h}^T\begin{bmatrix}
        \alpha R_c & \\ & 1
        \end{bmatrix}\begin{bmatrix}
        I \\
        \tau^T & 1
        \end{bmatrix}\hat{t}  + (\alpha - 1)\\
        \alpha&(\gamma - 1) + 1 < \hat{h}^TR_*'t,
        \end{aligned}
    \end{equation}
    since $R_*$ and $\alpha$ are learnable, the margin can be dynamic adjust without changing $h$ and $t$. Thus, we could say that STaR can handle complex relations.
    % \item \textbf{Non-Inversion}: If $R_c$ is not full rank, then $R_{*}$ is non-invertible and thus does not have inversion elements correspondingly.
\end{enumerate}
Based on the discussion above, we could conclude that STaR is capable to model all 6 patterns and handle complex relations.
\end{proof}
\section{Details of DURA}
\label{app:detail}
% The weight $w(t)$ is calculated as:
%  Adopt form \cite{dura}, we set $w_0 = 0.1$ for all models on WN18RR and RESCAL on YAGO3-10, while $w_0 = 0$ for other cases.

For a bilinear model $h^TRt$ in real value, the DURA regularization is:
\begin{align}
    \label{equ:dura}
    \|h\|^2_2 + \|Rt\|_2^2 + \|t\|_2^2 + \|h^TR\|_2^2.
\end{align}
Then, for STaR, we have:
\begin{equation}
    \begin{aligned}
    &\|h\|^2_2 + \|Rt\|_2^2 + \|t\|_2^2 + \|h^TR\|_2^2 \\
    =&\|\hat{h}\|^2_2 + \|R_*\hat{t}\|_2^2 + \|\hat{t}\|_2^2 + \|\hat{h}^TR_*\|_2^2 \\
    =&\|h\|^2_2 + \|t\|^2_2 + \|h^TR_c+\tau\|^2_2 + \|R_ct\|^2_2 + \tau^Tt + 4.
    \end{aligned}
\end{equation}
The emergence of constant $4$, which can be ignored in the optimization,  is because we use $\hat{h},\hat{t}$ having an extra dimension with constant 1.

\section{Details of $\psi$ and $\Psi$}
\label{app:matrix}
\begin{figure}[ht]
    \centering
    \includegraphics[width=0.4\textwidth]{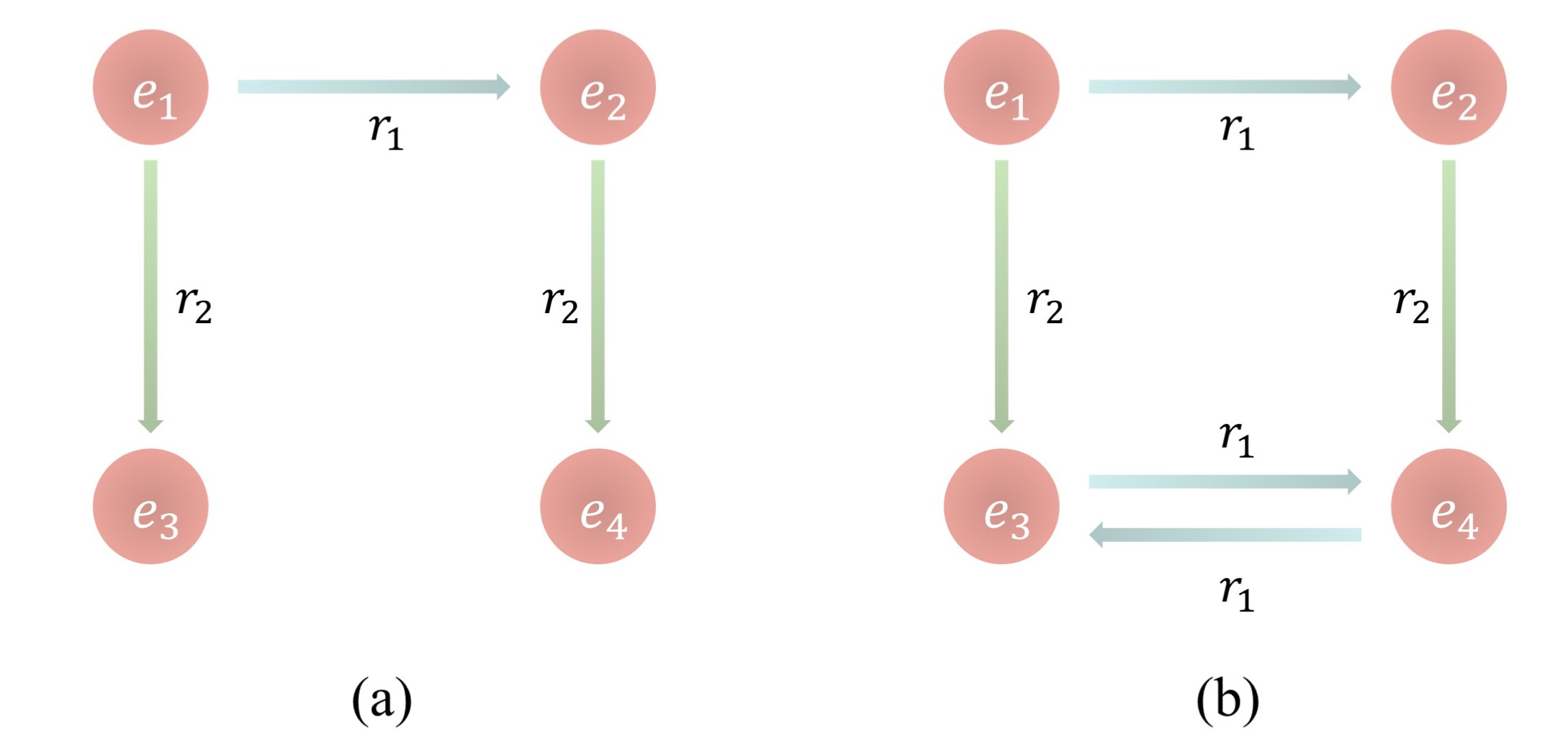}
    \caption{Toy examples demonstrate how to count $\#\text{triple}_{ij}$}
    \label{fig:order_example}
\end{figure}
 For each possible relation pair $(r_i, r_j) \in (\mathcal{R}, \mathcal{R})$, we count its corresponding triple in the training set $\mathcal{T}_{train}$ as $\#\text{triple}_{ij}$ and $\#\text{triple}_{ji}$. For Instance, in the Figure \ref{fig:order_example}, $\#\text{triple}_{12} = 1$ and  $\#\text{triple}_{21} = 0$ in the left hand example while $\#\text{triple}_{12} = 1$ and  $\#\text{triple}_{21} = 2$ in the right one. Then, we define the imbalance ratio of a relation pair $\psi_{ij}$ as:

\begin{align}
    \label{equ:pair ratio}
    \psi_{ij} = 2\cdot\frac{\max{\{\#\text{triple}_{ij},\#\text{triple}_{ji}\}}}{\#\text{triple}_{ij} + \#\text{triple}_{ji}} - 1.
\end{align}

Meanwhile, we treat a pair as \textit{both} if $\#\text{triple}_{ij} > 0$ and $\#\text{triple}_{ji} > 0$, and \textit{single} if only one of them greater than $0$. Based on that, we count the triples of \textit{both} and \textit{single} as $\#\text{triple}_{both}$ and $\#\text{triple}_{single}$ respectively. Thus, we define a similar matrix $\Psi$ for the imbalance ratio of train set:
\begin{equation}
    \Psi = \frac{\#\text{triple}_{single}}{\#\text{triple}_{both} + \#\text{triple}_{single}}.
\end{equation}

\end{document}